\documentclass[preprint,12pt]{elsarticle}

\usepackage{soul, color}
\usepackage{graphicx}
\usepackage{adjustbox}
\usepackage{booktabs,makecell}
\usepackage{makecell}
\usepackage{tabularx}
\usepackage{threeparttable, tablefootnote}
\usepackage{amsmath}

\usepackage{newunicodechar}

\usepackage[colorlinks,allcolors=black]{hyperref}
\usepackage{hyperref}
\hypersetup{colorlinks,allcolors=black}

\usepackage{graphicx}
\usepackage{multirow}
\usepackage[dvipsnames]{xcolor}
\usepackage{soul}
\usepackage{color}
\usepackage{url}
\usepackage{float}
\usepackage{boldline}
\usepackage{xcolor,colortbl}
\usepackage{tikz}

\usepackage{pifont}

\usepackage{multirow}
\usepackage{graphicx}
\usepackage{amssymb}

\usepackage{soul}
\usepackage[normalem]{ulem}
\usepackage{xcolor}

\usepackage{xspace}
\usepackage{tabularx}

\journal{}

\begin{document}

\begin{frontmatter}


\title{Exploring Kolmogorov-Arnold Network Expansions in Vision Transformers for Mitigating Catastrophic Forgetting in Continual Learning}



\author{Zahid Ullah$^{1}$,  Jihie Kim$^{1,*}$}

\address{%
$^{1}$ \quad Department of Computer Science and Artificial Intelligence, Dongguk University, Seoul 04620, Republic of Korea \\   } 

\begin{abstract}
Continual learning (CL), the ability of a model to learn new tasks without forgetting previously acquired knowledge, remains a critical challenge in artificial intelligence, particularly for vision transformers (ViTs) utilizing Multilayer Perceptrons (MLPs) for global representation learning. Catastrophic forgetting, where new information overwrites prior knowledge, is especially problematic in these models. This research proposes replacing MLPs in ViTs with Kolmogorov-Arnold Network (KANs) to address this issue. KANs leverage local plasticity through spline-based activations, ensuring that only a subset of parameters is updated per sample, thereby preserving previously learned knowledge. The study investigates the efficacy of KAN-based ViTs in CL scenarios across benchmark datasets (MNIST, CIFAR100), focusing on their ability to retain accuracy on earlier tasks while adapting to new ones. Experimental results demonstrate that KAN-based ViTs significantly mitigate catastrophic forgetting, outperforming traditional MLP-based ViTs in knowledge retention and task adaptation. This novel integration of KANs into ViTs represents a promising step toward more robust and adaptable models for dynamic environments. 

The code implementation of our approach will be openly shared after publication to promote reproducibility and facilitate further research. Code Link: \url{https://github.com/Zahid672/KAN-CL-ViT-main/tree/main?tab=readme-ov-file}.
\end{abstract}

\end{frontmatter}


\section{Introduction}
\label{intro}
\begin{figure*}[!ht]
     \centering
     \includegraphics[width=1\textwidth]{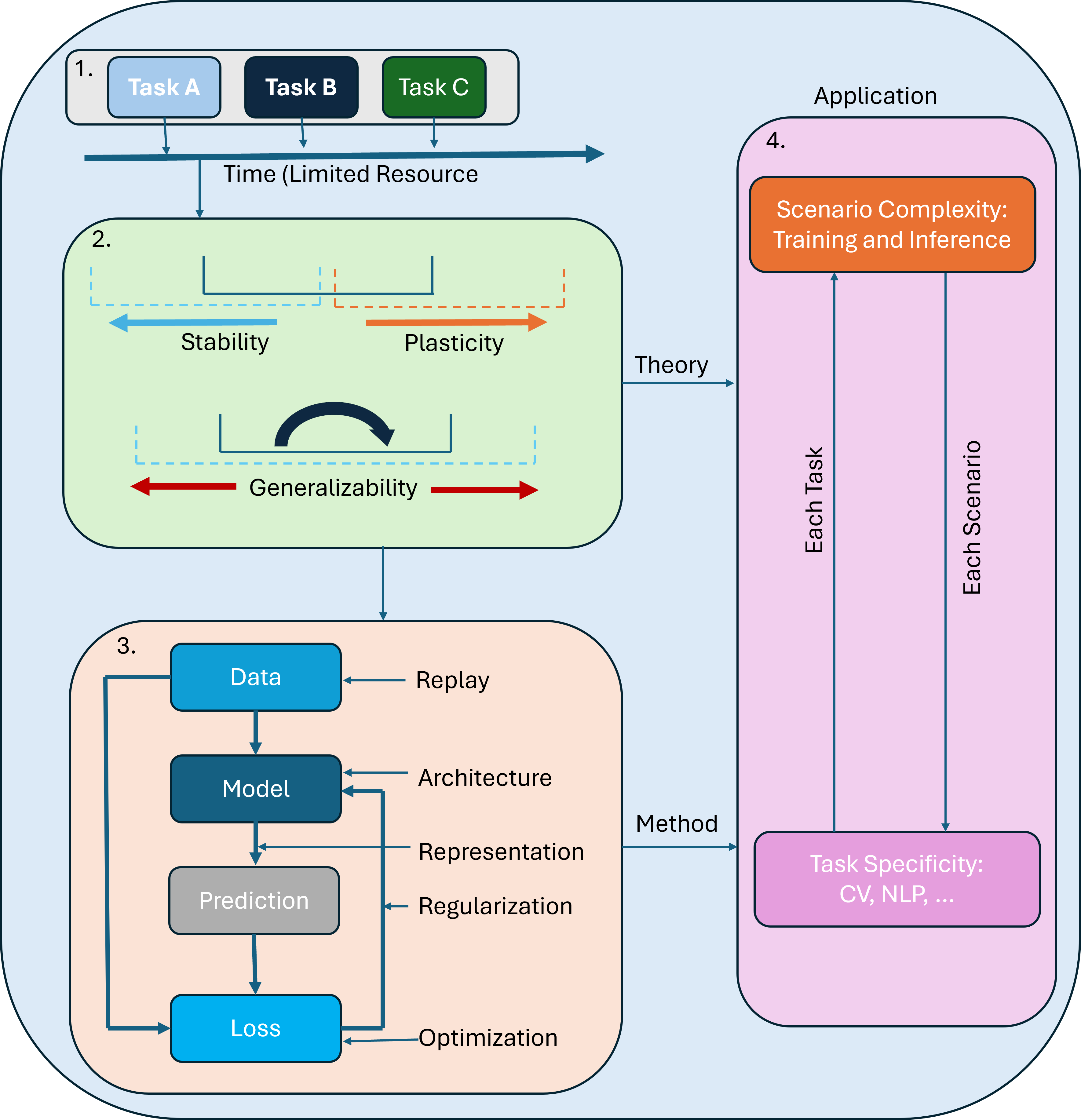}
     \caption{A conceptual framework for CL. 1, involves adapting to a incremental tasks where data distributions change over time. 2, An effective approach must balance stability (shown by the blue arrow) and plasticity (orange arrow), while also maintaining strong generalization both within individual tasks (black arrow) and across different tasks (red arrow). 3, To meet the goals of CL, leading methods have focused on different machine learning components. 4, In real-world applications, CL is tailored to overcome specific challenges such as the complexity of scenarios and the unique requirements of each task. }
     \label{CLC}
 \end{figure*}

\begin{figure*}[!ht]
     \centering
     \includegraphics[width=1\textwidth]{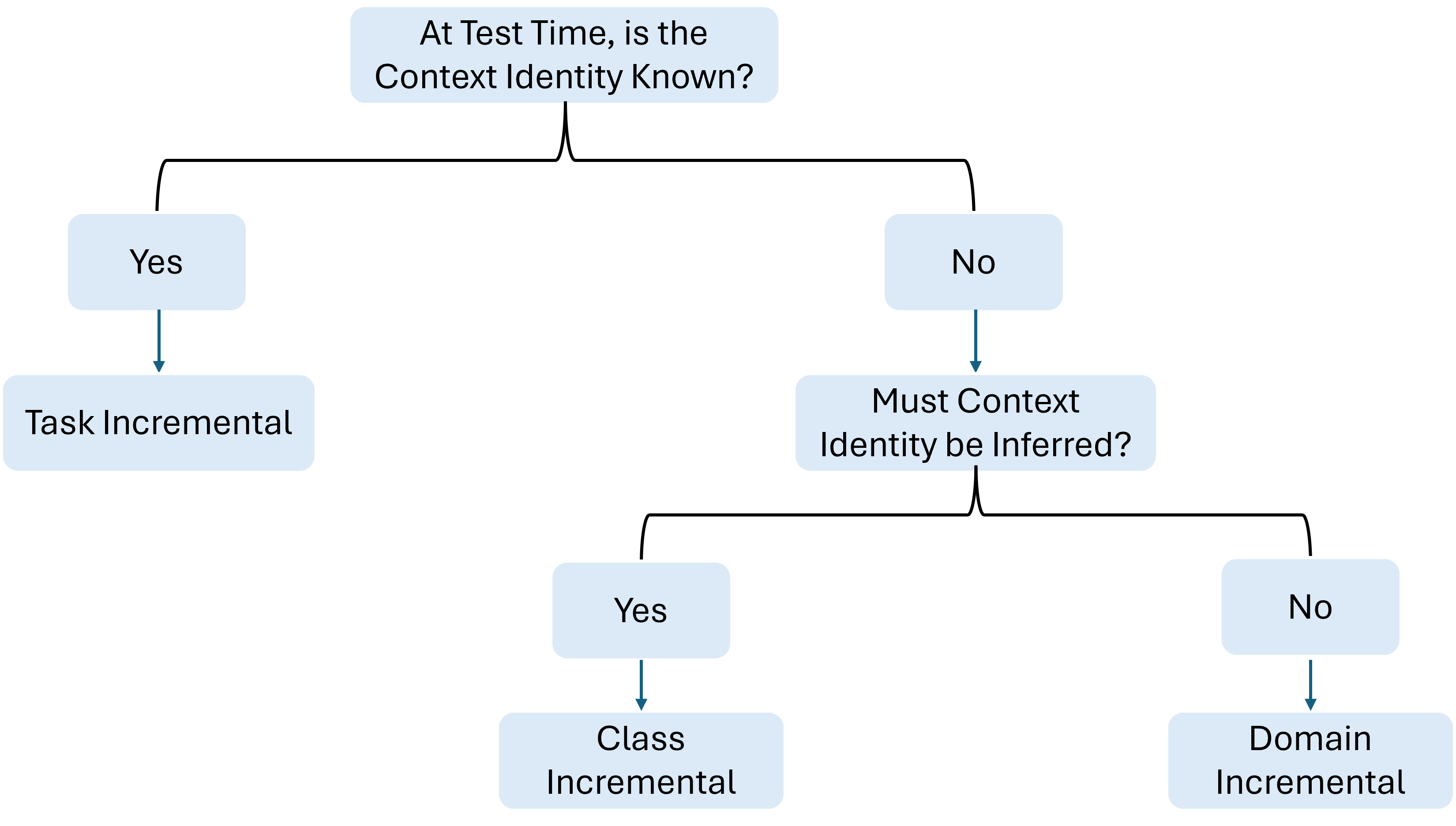}
     \caption{Decision tree for the three CL scenarios. }
     \label{DT}
 \end{figure*}

Incremental learning \cite{wang2024comprehensive,van2022three}, also referred to as continual learning (CL), addresses the critical challenge of enabling machine learning models to adapt to new data and tasks sequentially without forgetting previously acquired knowledge (See Fig. \ref{CLC}). This capability is essential for real-world applications where data distributions evolve over time or where it is impractical to retrain models from scratch with the entire dataset each time new information becomes available \cite{gupta2025personalized}.  CL algorithms strive to build versatile AI agents by learning from a continuous sequence of tasks, thereby contributing to the development of generally intelligent systems \cite{lesort2021understanding,lesort2020continual,xu2025multi}. However, achieving stable and robust incremental learning is difficult because neural networks often suffer from catastrophic forgetting, where the acquisition of new information leads to a drastic decline in performance on old tasks \cite{parisi2019continual}. To mitigate catastrophic forgetting, various strategies have been developed, which can be broadly categorized into three main scenarios: task-incremental learning, domain-incremental learning, and class-incremental learning as shown in Fig. \ref{DT}.

Task-incremental learning assumes that task identities are known at both training and test times, allowing the model to utilize this information to maintain task-specific knowledge \cite{parisi2019continual,aljundi2017expert}. In this scenario, the model incrementally learns a set of clearly distinguishable tasks, where the boundaries between tasks are well-defined \cite{de2021continual}. At inference, the model is provided with the task identity, allowing it to select the appropriate parameters or modules for that specific task. One common approach in task-incremental learning involves using separate model components or parameters for each task, such as dynamically expandable networks. These methods allocate new capacity for each task, preventing interference between different tasks and mitigating catastrophic forgetting. However, these approaches can suffer from scalability issues as the number of tasks increases, requiring significant memory and computational resources. Another strategy involves using task-specific modulation techniques, where task embeddings are used to modulate the activations or parameters of a shared network.  These modulation techniques allow the model to adapt its behavior to each task while maintaining a shared representation across all tasks. Task-incremental learning provides a simplified setting for studying CL, as the task identity provides a strong signal for guiding the learning process and preventing interference between tasks, but in realistic scenarios task boundaries are often ambiguous or unknown, making these methods impractical.

Domain-incremental learning \cite{ke2021classic,mirza2022efficient} presents a more challenging scenario where the model learns the same problem but in different contexts, leading to shifts in the data distribution over time. In this setting, the task identity is not explicitly provided, but the model must adapt to changing input distributions while preserving previously learned knowledge \cite{parisi2019continual}. Domain-incremental learning is relevant in many real-world applications where the environment or input data characteristics change over time, such as in robotics, autonomous driving, and financial modeling. A key challenge in domain-incremental learning is to detect and adapt to these distribution shifts without forgetting previously learned domains. One popular approach involves using domain adaptation techniques to align the feature distributions of different domains. These methods aim to learn domain-invariant representations that are robust to changes in the input distribution, allowing the model to generalize across different domains. Another strategy involves using techniques like replay, where data from previous domains is stored and replayed during training on new domains. By replaying old data, the model can consolidate its knowledge and prevent catastrophic forgetting.  However, replay-based methods can be limited by memory constraints and privacy concerns. Alternatively, regularization-based methods add a penalty term to the loss function to prevent the model's parameters from deviating too much from their previous values. The ability to discern which features vary across source and target datasets is crucial for reducing domain discrepancy and ensuring accurate knowledge transfer \cite{uguroglu2011feature,mounsaveng2019adversarial}.

Class-incremental learning \cite{li2024multi,tao2020few,masana2022class,cossu2022class} poses the most challenging scenario, where the model must incrementally learn to distinguish between a growing number of classes without revisiting data from previous classes. This setting is particularly relevant in applications such as image recognition, object detection, and natural language processing, where the set of possible categories or concepts can expand over time. Class-incremental learning is challenging because the model must learn new classes without forgetting previously learned classes, and without having access to the data from previous classes. This can lead to a significant bias towards the new classes, resulting in a decline in performance on old classes, or catastrophic forgetting. One common approach involves using techniques like knowledge distillation, where the model learns to mimic the output of a previous model trained on the old classes. This helps to preserve the knowledge learned from the old classes while adapting to the new classes. Another strategy involves using techniques like exemplar replay, where a small subset of data from previous classes is stored and replayed during training on new classes. However, exemplar replay methods can be limited by memory constraints and the need to carefully select representative exemplars.  Generative replay, where a generative model trained on previous tasks produces synthetic data that is interleaved with data from the current task, is another alternative \cite{masip2023continual}. When combined with methods like parameter regularization, data augmentation, or label smoothing, activation regularization techniques can improve model robustness \cite{liu2022towards}. Data augmentation, which involves applying transformations to the training data, can also improve the generalization ability of the model and prevent overfitting \cite{song2022learning,mounsaveng2019adversarial}.

Among these three approaches, task-based incremental learning presents a distinct paradigm within the realm of CL, offering notable advantages over other incremental learning methodologies by explicitly delineating the learning process into discrete tasks. This structured approach facilitates a more controlled and interpretable learning trajectory, mitigating the detrimental effects of catastrophic forgetting that often plague traditional machine learning models when exposed to sequential data streams \cite{gupta2025personalized}. The explicit task boundaries enable the implementation of targeted strategies to preserve previously acquired knowledge, such as rehearsal-based methods that replay representative samples from past tasks or regularization techniques that constrain the model's parameter drift, preventing it from deviating too far from previously learned representations. Furthermore, task-based incremental learning aligns well with real-world scenarios where tasks are naturally segmented, allowing for seamless integration of new skills or knowledge without disrupting existing capabilities \cite{de2021continual}. By leveraging the locality of splines, Kolmogorov-Arnold Network (KANs) can avoid catastrophic forgetting. This paradigm allows for more flexible and expressive function approximations. 

The clear task separation inherent in task-based incremental learning allows for the development of specialized learning strategies tailored to the unique characteristics of each task. This adaptability enhances the overall learning efficiency and effectiveness, enabling the model to acquire new knowledge more rapidly and accurately. For instance, a model might employ a more aggressive learning rate for tasks with abundant data and a conservative learning rate for tasks with limited data to prevent overfitting. Moreover, task-based incremental learning facilitates the evaluation and comparison of different learning algorithms in a standardized manner, providing a clear benchmark for assessing their performance on specific tasks and their ability to retain knowledge across multiple tasks. The explicit task boundaries also enable the use of task-specific meta-learning techniques, where the model learns to learn new tasks more efficiently based on its experience with previous tasks \cite{lesort2021understanding}. When a network learns a new task, it modifies its weights to minimize the error for that specific task. These weight updates can significantly alter the knowledge acquired from previous tasks.

KANs offer a compelling approach to function approximation, underpinned by the Kolmogorov-Arnold representation theorem, which posits that any multivariate continuous function can be represented as a composition of univariate functions. This theoretical robustness makes KAN particularly suited for complex learning scenarios. In the context of CL, where models are required to sequentially acquire knowledge across multiple tasks without forgetting previous ones, KAN's inherent design aligns well with the need for both compact representations and knowledge retention.

Furthermore, CL presents significant challenges, particularly in mitigating catastrophic forgetting. Traditional architectures like multilayer perceptrons (MLPs) often struggle to balance the trade-off between learning new tasks (plasticity) and retaining old ones (stability). Investigating KAN allows us to explore its potential to improve stability, maintain generalization, and provide more robust solutions for CL tasks, particularly in the image classification domain.

Moreover, vision transformers (ViTs), inspired by the success of transformers in natural language processing, have revolutionized computer vision \cite{dahan2022surface} by employing self-attention mechanisms \cite{grosz2023afr} to achieve state-of-the-art performance across various tasks. The ability of transformers to model long-range dependencies and capture global context has proven particularly advantageous in image recognition, object detection \cite{kheddar2025transformers}, and semantic segmentation \cite{dosovitskiy2020image,wu2025powerful}. However, the inherent architecture of ViTs, particularly the reliance on MLPs within each transformer block, presents a significant challenge, such as catastrophic forgetting. Catastrophic forgetting refers to the phenomenon where a neural network, after being trained on a sequence of tasks, abruptly loses its ability to perform well on previously learned tasks when trained on a new task. This limitation hinders the deployment of ViTs in real-world CL scenarios where models must adapt to new data without forgetting previously acquired knowledge.

The core problem of catastrophic forgetting in MLPs stems from their global representation learning approach, where each neuron's weights are updated based on every input sample, thus remodeling the entire region after seeing new data samples. This contrasts sharply with the hypothesis presented by Liu et al. \cite{liu2024kan} which posits that KANs can mitigate catastrophic forgetting due to the local plasticity afforded by their spline-based activations. The idea is that each sample predominantly influences only a small subset of spline coefficients, leaving the majority of the network parameters untouched, thus preserving previously learned knowledge. This research is grounded in the assertion that KANs' local plasticity offers a pathway to alleviate catastrophic forgetting in ViTs.

 To investigate this claim, this work explores the replacement of MLPs in ViTs with KANs. This modification aims to enhance the ability of ViTs to retain knowledge across multiple tasks, reducing the impact of catastrophic forgetting. Furthermore, this study aims to enhance the ability of ViTs to retain knowledge across multiple tasks, reducing the impact of catastrophic forgetting, and provide a comprehensive analysis of  adam optimizer and its effects on KAN-based ViTs. The objective of this work is to rigorously investigate the performance of KAN-based ViTs in CL settings, focusing on their ability to maintain accuracy on previously learned tasks while adapting to new information.  This will also involve experimenting with adam optimizers, followed by conducting experiments to assess the KAN-based model's ability to retain knowledge across multiple tasks. By exploring these questions, this research seeks to provide insights into the potential of KANs as a solution to the challenge of catastrophic forgetting in ViTs, paving the way for more robust and adaptable vision models.

Specifically, this study aims to evaluate the effectiveness of substituting MLPs with KANs in ViTs to alleviate catastrophic forgetting and improve the models' overall learning capabilities. This objective will be achieved by comparing the performance of KAN-based ViTs with traditional MLP-based ViTs in various CL scenarios. The outcomes include improved CL, which enhances the ability of ViTs to retain knowledge across multiple tasks, and optimization insights, which provide a comprehensive analysis of various optimizers and their effects on KAN-based ViTs.  This research is significant because it offers a potential solution to the critical problem of catastrophic forgetting in ViTs, enhancing their applicability in dynamic and evolving environments. This capability is crucial for real-world applications where models must continuously adapt to new data without losing previously acquired knowledge. We achieved improved accuracy with KAN-based ViTs, surpassing traditional MLP-based ViTs by leveraging learnable activation functions on edges. The contributions of this study are stated below:
\begin{itemize}
    \item The replacement of traditional MLP layers in ViTs with KANs to leverage their local plasticity and mitigate catastrophic forgetting.

    \item In-depth exploration and evaluation of different KAN configurations, shedding light on their impact on performance and scalability.

\item Demonstration of superior knowledge retention and task adaptation capabilities of KAN-based ViTs in CL scenarios, validated through experiments on benchmark datasets like MNIST and CIFAR100.

\item Development of a robust experimental framework simulating real-world sequential learning scenarios, enabling rigorous evaluation of catastrophic forgetting and knowledge retention in ViTs.
\end{itemize}

The remaining paper is organized as follows: Section \ref{related} presents a comprehensive review of background knowledge to contextualize our research. Next, in Section \ref{pm} we detail our proposed methodology, including the dataset and the main architecture. We then describe our experimental setup and implementation details in section \ref{experimental}. Section \ref{Results} consists of the results and discussion. Finally, in section \ref{conclusion}, we present the conclusion and future work.

\section{Background Knowledge and Motivation}\label{related}
\subsection{Continual Learning and Its Importance}
CL is a critical paradigm in machine learning that focuses on training models to learn and adapt sequentially to multiple tasks over time. Unlike traditional machine learning approaches that train models in isolation with static datasets, CL aims to mimic human-like learning, where knowledge is incrementally acquired and retained. This capability is essential for building adaptive systems capable of functioning in dynamic environments, such as autonomous agents, robotics, and real-world applications where data arrives in streams or evolves over time. 
The primary objective of CL is to ensure that models retain knowledge of previously learned tasks while effectively acquiring new knowledge. However, achieving this balance is challenging, particularly in neural networks and other machine learning models, due to inherent limitations in how they process and store information.

\subsection{Catastrophic Forgetting: The Core Challenge}
One of the most significant hurdles in CL is catastrophic forgetting. This phenomenon occurs when a model trained on a sequence of tasks loses its ability to perform well on earlier tasks after being updated for new ones. Neural networks, in particular, are highly susceptible to catastrophic forgetting because they share parameters across all tasks. As these shared parameters are optimized for a new task, they may overwrite the knowledge crucial for previously learned tasks, leading to a sharp decline in performance. For instance, a neural network trained on Task A may achieve high accuracy, but after it is updated for Task B, its performance on Task A often deteriorates drastically. This challenge arises because traditional learning algorithms, such as stochastic gradient descent, lack mechanisms to protect task-specific knowledge during updates.

\subsection{Vision Transformers}
ViTs \cite{dosovitskiy2020image} have revolutionized computer vision tasks by leveraging self-attention mechanisms to capture long-range dependencies within images. However, they face challenges such as catastrophic forgetting, where the model's performance on previously learned tasks deteriorates when trained on new sequential tasks. This issue limits the application of ViTs in dynamic environments requiring CL. Moreover, ViTs demand substantial training data due to their extensive number of parameters and lack of inductive bias, posing challenges in medical imaging where data is scarce due to annotation efforts and data protection regulations \cite{shen2023movit}. To address catastrophic forgetting, KANs have been proposed as replacements for MLPs within ViT architectures. KANs leverage local plasticity from spline-based activations, enabling better task retention and mitigating knowledge interference.

\subsection{Kolmogorov–Arnold Networks}
KANs, rooted in the Kolmogorov-Arnold representation theorem, represent a class of neural architectures designed to model any continuous multivariable function as a finite sum of continuous univariate functions \cite{kolmogorov1957representations}. This foundational principle sets KANs apart from traditional neural networks by enabling localized learning through univariate spline-based activation functions.

\subsubsection{KANs in Neural Network Architectures}
Unlike MLPs, which rely on fixed global activation functions, KANs employ learnable activation functions on edges, parameterized by B-splines. This unique property enhances their capacity for fine-tuned function approximation while preserving local information. The introduction of efficient variants such as EfficientKAN reduces computational overhead, making KANs viable for memory-constrained applications \cite{krizhevsky2009learning}. Recent studies have demonstrated KANs’ advantages in areas such as: 

\textbf{Fine-grained learning:} Their ability to dynamically adjust activation functions provides greater precision in feature representation.

\textbf{Catastrophic forgetting mitigation:} By leveraging the locality of splines, KANs restrict updates to specific regions of their parameter space, preserving information about previously learned tasks \cite{liu2024kan}.

\subsubsection{Kolmogorov-Arnold Network in Continual Learning}
\begin{figure*}[!ht]
     \centering
     \includegraphics[width=1\textwidth]{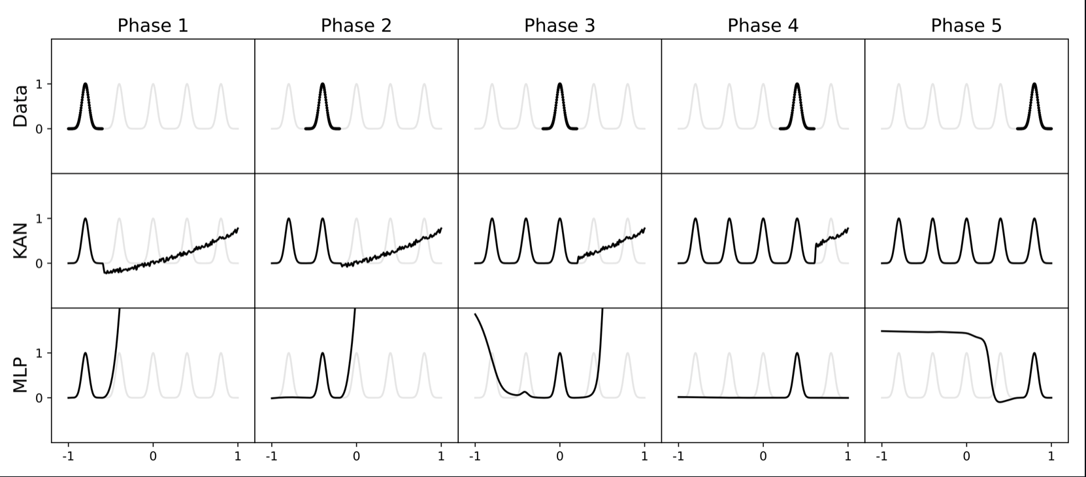}
     \caption{Catastrophic forgetting MLP vs KAN. The MLP waveforms become distorted as new phases are introduced—indicating catastrophic forgetting. Whereas, the KAN waveforms remain consistent across subsequent phases, showing better retention of knowledge across phases.
 }
     \label{catas}
 \end{figure*}

As illustrated in Fig. \ref{catas} KANs show potential for addressing catastrophic forgetting, a central challenge in CL. Unlike MLPs, whose global updates can distort prior knowledge, KANs retain information through local parameter updates. Liu et al. \cite{ramasesh2020anatomy} highlighted the efficacy of KANs in scenarios involving incremental and class-based CL. This research integrates KANs within ViTs architectures to explore their potential in CL. By replacing MLP layers with KANs, this study aims to evaluate improvements in task retention and learning efficiency across multiple tasks.

\subsection{Continual Learning and Incremental Learning} 
CL also known as lifelong learning, addresses the challenge of enabling machine learning models to learn continuously from a sequence of tasks while retaining knowledge from previous tasks. This paradigm is critical for applications that require adaptability and long-term learning without catastrophic forgetting—a phenomenon where learning new tasks overwrites knowledge of previous ones. Despite significant progress, achieving effective CL remains a challenging problem due to the conflicting requirements of stability and plasticity.  Incremental learning, a subset of CL, deals specifically with progressively expanding a model’s capability by introducing new tasks or classes over time. Both approaches are central to achieving adaptable and robust artificial intelligence.

\subsection{Task-Based Incremental Learning to Mitigate Catastrophic Forgetting}
In the context of CL, task-based incremental learning is a prominent approach where models are explicitly provided task identifiers during training and testing. This simplifies the learning process by focusing on task-specific adaptation while retaining prior knowledge. Unlike class-incremental or domain-incremental learning, task-based approaches benefit from clearer task boundaries, reducing ambiguity. For instance, the notable method in this scenario is, progressive neural networks \cite{rusu2016progressive}, which grow the architecture for new tasks while freezing previous parameters. However, this paradigm still struggles with knowledge retention and efficient transfer between tasks, especially when task boundaries are poorly defined or overlapping.

\subsection{Replay-Based Methods}

Replay-based methods have emerged as a cornerstone in mitigating catastrophic forgetting. These methods preserve knowledge from previous tasks by replaying data during training on new tasks. Two primary categories of replay-based methods are:

\textbf{Experience Replay:} This approach involves storing a subset of data from previous tasks in a memory buffer and interleaving this data with the current task during training. Techniques like reservoir sampling and prioritized replay optimize memory usage, enabling efficient performance even with constrained resources. For instance, Alvise et al. \cite{rebuffi2017icarl} maintain exemplars of past tasks to augment training data. Whereas, Lopez et al. \cite{lopez2017gradient} ensure gradients for new tasks do not interfere with old tasks by projecting them orthogonally.

\textbf{Generative Replay:} Generative models, such as Variational Autoencoders and Generative Adversarial Networks, are trained to synthesize data from previous tasks. This eliminates the need to store raw data, addressing privacy concerns and memory limitations. Despite these advantages, generative replay often suffers from fidelity issues in synthesized data and high computational demands. Specifically, replay-based methods are particularly well-suited for task-based incremental learning as they align with the need to maintain task-specific information while enabling incremental updates. However, replay strategies face challenges in scalability and balancing memory constraints with replay effectiveness.

\subsubsection{Comparative Analysis}
Replay-based methods are often contrasted with other CL strategies. Regularization-based methods, such as Elastic Weight Consolidation and Synaptic Intelligence, penalize changes to parameters critical to previous tasks, thereby reducing forgetting. Architectural approaches, like progressive networks, allocate separate modules for each task, avoiding interference but at the cost of scalability. Replay-based methods strike a balance between these extremes by leveraging prior task data to reinforce learning while maintaining flexibility for new tasks.

\subsubsection{Existing Gaps and Motivation}
Despite their success, replay-based approaches face limitations, including memory efficiency, effective task representation, and robustness to distribution shifts. Additionally, task-based incremental learning scenarios often lack comprehensive exploration of how replay-based strategies can enhance knowledge transfer across tasks. Addressing these gaps requires innovative mechanisms to optimize memory usage, improve generative data quality, and integrate task-specific insights into replay dynamics.

This work aims to advance the state of task-based incremental learning by proposing novel replay-based mechanisms tailored to task-specific requirements. By addressing the aforementioned challenges, this study contributes to the broader understanding and application of CL in dynamic environments. The integration of KAN into the ViT architecture represents an innovative attempt to combine the theoretical strengths of KAN with the practical successes of ViT. KAN's ability to approximate functions efficiently and its potential for better knowledge retention complement ViT's superior feature extraction and scalability. This hybrid approach seeks to answer several key questions:
\begin{itemize}
    \item Can KAN enhance ViT's capacity for CL by mitigating forgetting while maintaining learning efficiency?

\item Does the combination provide a synergistic effect that improves overall task performance, especially in challenging scenarios with complex datasets like CIFAR100?
\end{itemize}

This research is motivated by the desire to develop architectures that not only achieve high accuracy but also exhibit resilience and adaptability in CL environments, paving the way for more advanced solutions in lifelong learning systems.

\subsection{Summary}
KAN have shown promise in image classification tasks due to their foundation in the Kolmogorov-Arnold representation theorem, which states that any continuous multivariate function can be represented as a superposition of univariate functions. This mathematical framework enables KAN to approximate complex mappings with high efficiency and compactness, making them particularly suitable for image classification, where intricate patterns and relationships need to be captured. By leveraging this ability, KAN can efficiently model non-linear relationships in image data, leading to accurate and robust classification outcomes. Moreover, their structure may provide advantages in terms of adaptability and knowledge retention, which are critical in dynamic tasks like CL. In this context, KAN can serve as a standalone classifier or be integrated into more complex architectures, such as ViTs, to enhance feature extraction and classification performance. The unique theoretical strengths of KAN, combined with its practical applicability, make it a valuable tool for tackling challenges in modern image classification scenarios. The central hypothesis of this research is that replacing MLPs in ViTs with KANs will lead to a reduction in catastrophic forgetting and an enhancement of the overall learning capabilities of the model in CL scenarios. For instance, Liu et. al. \cite{liu2024kan} posit that KANs exhibit local plasticity and can avoid catastrophic forgetting by leveraging the locality of splines, highlighting that spline bases are local, so a sample will only affect a few nearby spline coefficients, leaving far-away coefficients intact, contrasting with MLPs that remodel the whole region after seeing new data samples. This research aims to empirically validate this claim and explore the effectiveness of KANs in various CL settings. ViTs represents a paradigm shift in computer vision, moving away from the traditional dominance of Convolutional Neural Networks \cite{mao2022towards}. Unlike ConvNets, which rely on convolutional layers to extract features, ViTs divide an image into patches and treat these patches as tokens, similar to words in a sentence, feeding them into a Transformer architecture \cite{aldahdooh2021reveal}. This approach allows ViTs to capture global relationships between image regions through self-attention mechanisms.

\section{Materials and Methods} 
\label{pm}


\subsection{Models and Variants}
\subsubsection{Overview of MLP, KAN, and KAN-ViT Architectures}
The study of MLP, KAN, and KAN-ViT architectures in image classification tasks, particularly in a CL context, highlights the distinctive characteristics and functionalities of these models. Each architecture brings unique theoretical and practical strengths to the domain of deep learning, offering diverse approaches to solving complex classification problems.

\subsubsection{Multi-Layer Perceptrons}
MLPs are one of the most fundamental architectures in deep learning, consisting of fully connected layers that transform input data through a series of weighted sums and nonlinear activation functions. In the context of image classification, MLPs process flattened image data, learning to map pixel values to corresponding class labels. The main advantages of MLP are it's simplicity, ease of implementation, and theoretical foundations. However, MLPs lack spatial awareness, making them less effective for high-dimensional image data where local patterns and spatial hierarchies are crucial. Despite these limitations, MLPs provide a baseline for evaluating advanced architectures, making them a critical component in CL experiments.

\subsubsection{Kolmogorov-Arnold Network}
KANs are inspired by the Kolmogorov-Arnold representation theorem, which states that any multivariate continuous function can be decomposed into univariate functions and a sum. This mathematical foundation enables KANs to approximate complex mappings with high precision, even in high-dimensional spaces. The KAN equation is as follows:
\begin{equation}
    f(\mathbf{x})=f\left(x_1, \cdots, x_n\right)=\sum_{q=1}^{2 n+1} \Phi_q\left(\sum_{p=1}^n \phi_{q, p}\left(x_p\right)\right),
\end{equation}

Here, $\varphi_{q, p}\left(x_p\right)$ are the spline functions, and $\Phi_q$ represent the transformations.

The activation function $\phi(x)$ is defined as follows:
\begin{equation}
    \phi(x)=w_b b(x)+w_s \operatorname{spline}(x),
\end{equation}
where $w$ represents a weight, $b(x)$ is the basis function (implemented as silu$(x))$, and spline$(x)$ is the spline function. 

The basis function $b(x)$ is defined as:
\begin{equation}
    b(x)=\operatorname{silu}(x)=x /\left(1+e^{-x}\right)
\end{equation}

The spline function spline$(x)$ is:
\begin{equation}
    \operatorname{spline}(x)=\sum_i c_i B_i(x)
\end{equation}

\textbf{Structure:} KANs utilize a hierarchical design, where univariate functions model specific input dimensions, and their interactions are synthesized through aggregation layers. This design allows for efficient representation and generalization across diverse tasks.

\textbf{Strengths in CL:} KANs are particularly advantageous in CL due to their ability to preserve learned representations. Their modular design helps mitigate catastrophic forgetting by isolating specific task representations. KANs outperform traditional MLPs in scenarios where preserving prior knowledge while learning new tasks is critical, as demonstrated in their performance on incremental classification tasks.

\subsubsection{Kolmogorov-Arnold Network-ViT}
The KAN-ViT architecture combines the theoretical strengths of KAN with the powerful feature extraction capabilities of ViTs. ViTs utilize self-attention mechanisms to capture global dependencies in image data, enabling them to model relationships between image patches effectively. In KAN-ViT, the KAN framework is embedded within the ViT architecture to enhance its adaptability and efficiency in learning diverse tasks. KAN modules replace or augment MLP layers in the transformer blocks, improving the model’s robustness in CL scenarios. The main advantage is KAN’s hierarchical function approximation complements the ViT’s attention mechanisms, resulting in richer feature representations. Also, KAN-ViT retains knowledge better than standard ViTs in sequential tasks, especially in the early stages of incremental learning. While KAN-ViT demonstrates an initial improvement in incremental accuracy compared to standalone MLP or ViT architectures, its performance converges with ViTs in later stages, highlighting areas for further optimization.

In summary, each of these architectures—MLP, KAN, and KAN-ViT—offers distinct advantages for image classification tasks. MLPs serve as a baseline model, while KAN provides a theoretically grounded approach for efficient function approximation and knowledge retention. KAN-ViT leverages the strengths of both KAN and ViTs, offering a hybrid solution that excels in early CL stages. Together, these architectures represent a spectrum of strategies for tackling the challenges of image classification in dynamic and evolving learning environments. Integrating KANs into the ViTs framework involves replacing the standard MLPs layers in ViT with KAN modules. This integration leverages the theoretical advantages of KAN's hierarchical function approximation while maintaining the powerful global feature extraction capabilities of ViT. Below is a detailed explanation of the integration process.

\subsubsection{Choosing a Standard Pre-trained Vision Transformer Model}
The process begins with selecting a standard ViTs model.. This model has demonstrated exceptional performance in image classification tasks due to their ability to process input images as sequences of patch embeddings and capture global relationships using self-attention mechanisms. The chosen pre-trained model provides a robust foundation for integrating KAN, ensuring that the self-attention mechanisms remain intact while enhancing the model's non-linear transformation capabilities.

\subsubsection{Replacement of MLP Layers with Kolmogorov-Arnold Network}
\begin{figure*}[!ht]
     \centering
     \includegraphics[width=1\textwidth]{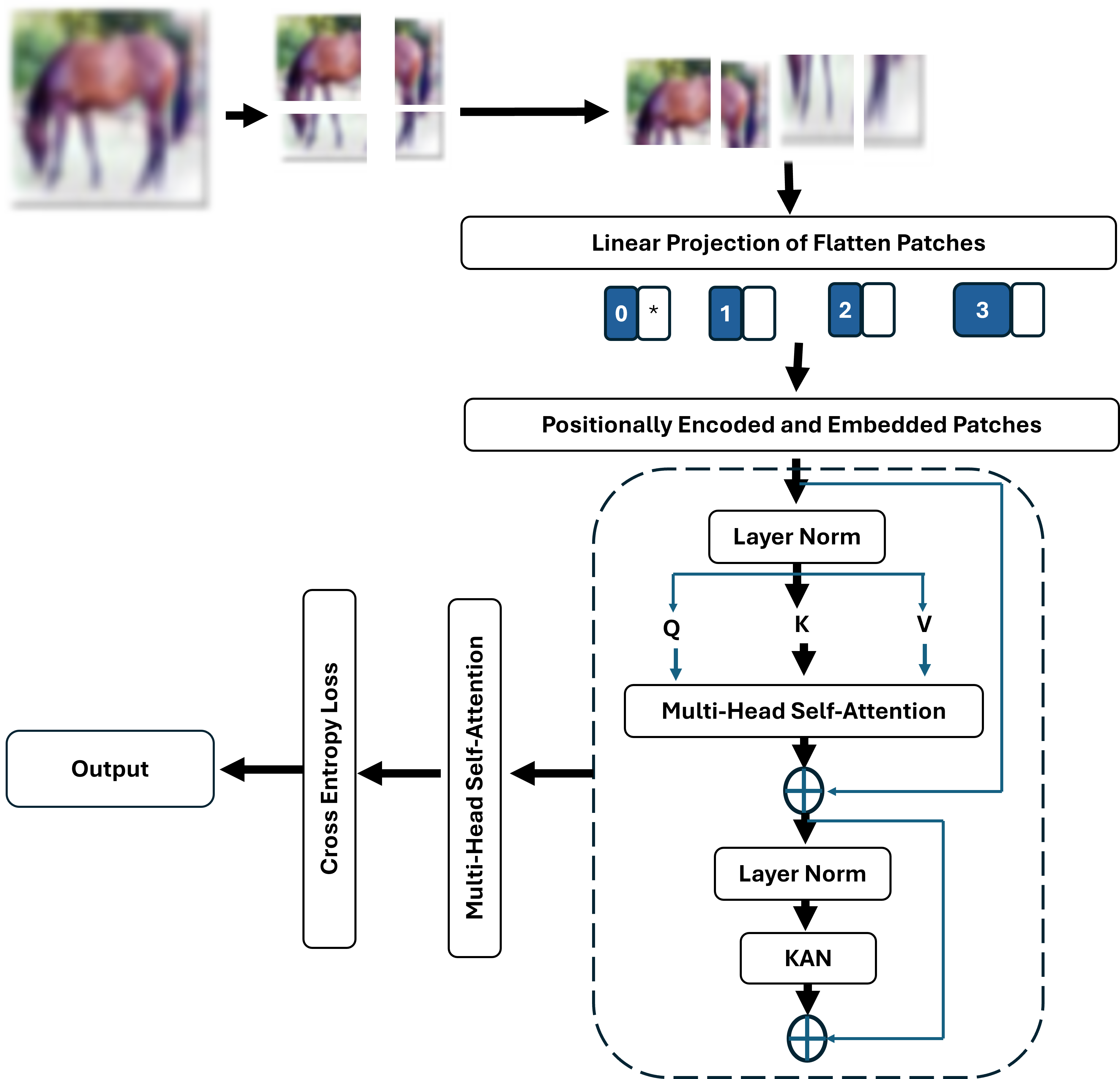}
     \caption{Our proposed methodology replaces the MLP layers in the selected ViT model with KANs. }
     \label{prop}
 \end{figure*}

In the standard ViT architecture, the MLP layers within each transformer block play a crucial role in processing the outputs of the self-attention mechanism. These MLP layers consist of fully connected layers that perform non-linear transformations and are vital for refining feature representations. KAN replaces these MLP layers with a modular architecture inspired by the Kolmogorov-Arnold theorem as shown in Fig. \ref{prop}. The KAN modules decompose complex multi-dimensional mappings into simpler univariate functions and their compositions. This hierarchical decomposition provides several advantages, such as, KAN's ability to approximate complex functions makes it highly suitable for learning intricate patterns in image data. In addition, KAN’s modular structure helps isolate representations, reducing interference between tasks in CL scenarios.

\subsubsection{Ensuring Seamless Integration of KAN Layers}
To ensure smooth integration, the KAN modules are designed to be compatible with the dimensions and computational requirements of the original MLP layers. The input and output dimensions of the KAN modules align with those of the replaced MLP layers, ensuring no disruption in the transformer block pipeline. The hierarchical structure of KAN is implemented to retain a similar computational footprint, making it scalable to large datasets and high-dimensional image tasks. The integration process may involve fine-tuning hyperparameters, adopting a specialized optimizer, or employing regularization techniques to ensure stable and efficient training of KAN layers within the ViT framework.

\subsubsection{Synergy between Kolmogorov and Vision Transformer}
The combination of KAN and ViT creates a synergistic effect:

\textbf{Self-Attention Complements KAN:} While ViT’s self-attention mechanism excels at modeling global dependencies between image patches, KAN enhances the non-linear transformations applied to these dependencies, enriching the overall feature representation.

\textbf{Modular and Scalable:} KAN's modular design integrates seamlessly into the sequential structure of ViT, ensuring adaptability to various datasets and tasks.

\textbf{Improved CL:} KAN’s capacity to isolate learned knowledge makes the hybrid model more resistant to catastrophic forgetting when compared to traditional MLPs.

The ViTs architecture has revolutionized the field of computer vision, demonstrating remarkable performance across various tasks \cite{shen2023movit}. However, despite their success, ViTs often grapple with the challenge of catastrophic forgetting, wherein the model's ability to retain previously learned knowledge diminishes upon exposure to new tasks. To address this limitation, this paper explores the integration of KANs as replacements for the traditional MLPs layers within a standard ViTs architecture, aiming to enhance knowledge retention and mitigate catastrophic forgetting. ViTs inherently capture long-range dependencies and enable parallel processing, yet lack inductive biases and efficiency benefits, facing significant computational and memory challenges that limit their real-world applicability \cite{zhang2025image}. The fundamental concept of ViTs involves partitioning an input image into a sequence of patches, which are then treated as tokens akin to words in natural language processing \cite{pereira2024review}. By replacing the MLP layers in ViT with KANs, it is hypothesized that the resulting architecture will exhibit superior performance in CL scenarios, demonstrating improved resistance to catastrophic forgetting while maintaining or enhancing accuracy on individual tasks. 

This research investigates the effects of replacing the standard MLPs in ViTs with KANs to address catastrophic forgetting and improve learning capabilities. KANs, with their inherent properties of local plasticity, offer a promising avenue for mitigating catastrophic forgetting by leveraging the locality of spline functions. KANs are based on the Kolmogorov-Arnold Representation theorem, which posits that any continuous function with multiple variables can be expressed as a finite sum of univariate function compositions. In this work, a standard pre-trained ViTs model is selected, and the MLP layers are replaced with KAN layers, ensuring a smooth integration within the ViT architecture. Unlike MLPs, which use fixed activation functions, KANs employ learnable activation functions on edges, enabling flexible and expressive function approximations. The inherent locality of spline bases in KANs ensures that each sample primarily affects only a few nearby spline coefficients, thereby minimizing the impact on distant regions and preserving previously learned knowledge. 

The integration of KANs into ViTs introduces a novel approach to CL, offering the potential to overcome the limitations of traditional MLP-based architectures. The objective is to enhance the ability of ViTs to retain knowledge across multiple tasks, reducing the impact of catastrophic forgetting, as well as providing a comprehensive analysis of various optimizers and their effects on KAN-based ViTs. Experimental results on image classification tasks reveal that KAN-based ViTs exhibit superior performance in CL scenarios, demonstrating improved resistance to catastrophic forgetting compared to traditional MLP-based ViTs.  The implementation of Adaptive Token Merging mitigates information loss through adaptive token reduction across layers and batches by adjusting layer-specific similarity thresholds \cite{lee2025lossless}. 

The choice of optimizers plays a crucial role in the training and convergence of KAN-based ViTs. The investigation delves into the effects of first-order optimizers on the performance of KAN-based ViTs and the associated resource requirements. First-order optimizers, such as ADAM, are widely used due to their computational efficiency and ease of implementation \cite{dosovitskiy2020image}. By splitting images into small patches and processing them to extract high-level features, ViTs employ an attention mechanism to effectively model the relationships between tokens. The exploration of novel methods aims to enhance the efficiency and performance of ViTs across a wide range of computer vision tasks, while retaining key knowledge and high performance \cite{shi2024faster}.

The architecture's resistance to catastrophic forgetting is rigorously measured and compared against traditional MLP-based ViTs. The findings provide valuable insights into the strengths and limitations of KAN-based ViTs, paving the way for future research in CL and adaptive neural networks. B-splines, which are piecewise polynomial functions used to create smooth and flexible curves, offer local control, meaning adjustments to one part of the spline do not affect distant regions. CL approaches such as replay, parameter regularization, functional regularization, optimization-based approaches, context-dependent processing, and template-based classification are employed to further mitigate catastrophic forgetting. This investigation ultimately seeks to provide a thorough understanding of the capabilities of KAN-based ViTs in CL settings, highlighting their potential for real-world applications where knowledge retention is paramount.

\subsection{Datasets and Task Splits}
To rigorously evaluate the efficacy and generalizability of our proposed methodology, we conducted a comprehensive suite of experiments utilizing two benchmark datasets widely recognized in the field of machine learning: MNIST \footnote{https://www.kaggle.com/datasets/hojjatk/mnist-dataset} and CIFAR100 \cite{krizhevsky2009learning} \footnote{https://www.kaggle.com/c/cifar100-image-classification/data}. These datasets were chosen due to their distinct characteristics and representational complexities, allowing for a thorough assessment of our approach under varying conditions \cite{sharma2018analysis}. The MNIST dataset, comprises a large collection of handwritten digits ranging from 0 to 9 \cite{basu2021handwritten}. It serves as a foundational dataset for image classification tasks, enabling researchers to prototype and evaluate novel algorithms with relative ease. CIFAR100, on the other hand, presents a more challenging scenario with its collection of 100 distinct object classes, each containing a diverse set of images. The use of publicly available datasets and benchmarks has been crucial to machine learning progress by allowing quantitative comparison of new methods \cite{picek2022danish}.

In the experiments conducted to evaluate the performance of MLPs and KANs, the MNIST and CIFAR100 datasets were strategically divided into multiple tasks to simulate a CL scenario. This setup was designed to mimic real-world environments where models must learn and adapt to sequentially presented data without access to previously seen information. By segmenting the datasets into distinct, non-overlapping tasks, the experiments aimed to test the model's ability to retain knowledge from earlier tasks while learning new ones, addressing the critical challenge of catastrophic forgetting.
\begin{figure*}[!ht]
     \centering
     \includegraphics[width=1\textwidth]{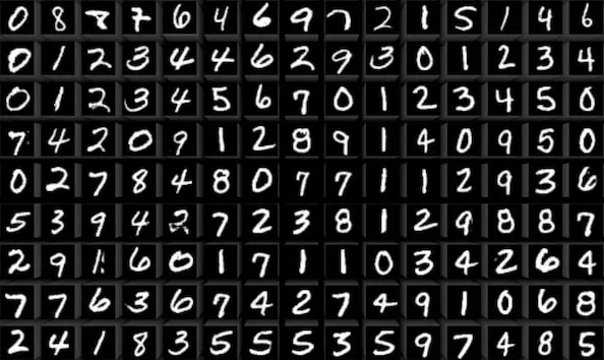}
     \caption{Images from MNIST dataset.} 
     \label{MDS}
 \end{figure*}

For the MNIST dataset, which contains 10 classes representing digits from 0 to 9 as shown in Fig. \ref{MDS} \footnote{https://datasets.activeloop.ai/docs/ml/datasets/mnist/}, the data was divided into 5 tasks, each comprising 2 classes. For example, the first task involved training the model on classes 0 and 1, the second task on classes 2 and 3, and so forth, until all 10 classes were covered. This sequential division ensured that each task introduced entirely new data, requiring the model to update its knowledge base while being unable to revisit the data from previous tasks. The training process was tailored to the incremental learning setup, with 7 epochs allocated for the first task to allow the model sufficient time to build a strong foundation. For the remaining tasks, the model was trained for 5 epochs each, testing its efficiency in integrating new information without over-relying on extended training.

\begin{figure*}[!ht]
     \centering
     \includegraphics[width=1\textwidth]{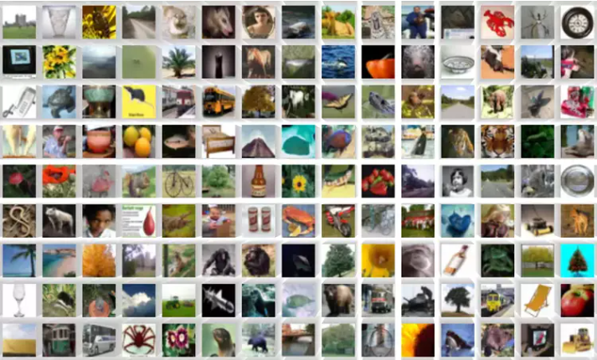}
     \caption{Images from CIFAR100 dataset.} 
     \label{CIFDS}
 \end{figure*}

The CIFAR100 dataset, a more complex and diverse dataset consisting of 100 classes, was divided into 10 tasks, with each task containing 10 distinct classes. Similar to the MNIST setup, the tasks were arranged sequentially, with the first task involving classes 0 to 9, the second task covering classes 10 to 19, and so on, until all 100 classes were utilized. This division introduced a greater challenge compared to MNIST, as the model had to handle a larger number of classes with more intricate visual features. The training schedule for CIFAR100 reflected its complexity: 25 epochs were dedicated to the first task to enable the model to establish a robust initial representation. For the subsequent tasks, the model was trained for 10 epochs per task, emphasizing rapid adaptation to new data while retaining prior knowledge. Images from CIFAR100 dataset is illustrated in Fig. \ref{CIFDS} \footnote{https://datasets.activeloop.ai/docs/ml/datasets/cifar-100-dataset/}.

By structuring the datasets in this way, the experiments effectively simulated the sequential nature of CL, where the model encounters new information in stages and must balance learning new tasks (plasticity) with preserving previously acquired knowledge (stability). The absence of access to prior task data heightened the challenge, requiring the model to rely on its internal mechanisms to mitigate catastrophic forgetting. This experimental design provided a controlled and rigorous environment to assess the potential of KAN and its integration with ViT for CL applications.

\subsection{Evaluation Metrics}
We have utilized various evaluation parameters to evaluate the performance of our proposed method.

\subsubsection{Last Task Accuracy}
We measure the overall performance of the model on the test set of all tasks, by calculating the average accuracy of the model on the test set of the classes that it has seen until task $T^b$. This metric is denoted by $A_K$.

\begin{equation}
LA=A_K \\
\end{equation}

\begin{equation}
A_b=\frac{1}{b} \sum_{i=1}^b a_{c, b}
\end{equation}

\subsubsection{Average Incremental Accuracy}
We measure the average performance of the model on all tasks, by computing the average accuracy of the model for each task, and then taking the average of these accuracies across all tasks.

\begin{equation}
    A I A=\frac{1}{K} \sum_{b=1}^K A_b 
\end{equation}

\subsubsection{Average Global Forgetting}
We quantify the model's ability to retain knowledge from previous tasks during the process of learning new tasks. This is achieved by calculating the difference in accuracy between the previous task and the current task. This metric is denoted by:

\begin{equation}
    F_G=A_{k-1}-A_k
\end{equation}

Furthermore, the average global forgetting is measured by:

\begin{equation}
    F_{A v g G}=\frac{1}{K} \sum_{k=1}^K\left(A_{k-1}-A_k\right)
\end{equation}

\subsubsection{Experimental Validation and Benefits}
Empirical results from experiments on datasets such as MNIST and CIFAR100 validate the effectiveness of KAN-ViT. These experiments demonstrate that the integration of KAN enhances the model’s performance in early stages of incremental learning by improving average incremental accuracy and knowledge retention. The hybrid model performs competitively in later stages, maintaining the advantages of both KAN’s theoretical strengths and ViT’s powerful attention mechanisms. Replacing the MLP layers in ViT with KAN modules transforms the architecture into a more robust and adaptive system. This integration not only preserves the strengths of ViT but also incorporates the function approximation and modular advantages of KAN, making the hybrid KAN-ViT model a compelling choice for complex classification tasks, particularly in CL scenarios.

\section{Experimental setup}
\label{experimental}
The experimental setup details are described in their corresponding subsections.

\subsection{Training Procedure}
The training epochs for Task 1 and the remaining tasks in the MNIST and CIFAR100 experiments were carefully designed to balance the model’s ability to learn the initial task thoroughly while maintaining computational efficiency for subsequent tasks. This division reflects the unique demands of CL, where the model must acquire knowledge incrementally while mitigating catastrophic forgetting of prior tasks. Below is a detailed explanation of the training strategy:

\subsubsection{Training Epochs for Task 1}
\textbf{1. MNIST Experiments}

\begin{itemize}
    \item \textbf{Epochs for Task 1 : 7}

The first task in the MNIST experiments is allocated 7 epochs, slightly more than the subsequent tasks. This additional training time allows the model to establish a solid foundation by thoroughly learning the features of the initial set of classes (2 classes in this case).

\item \textbf{Reasoning:} Since the model starts with no prior knowledge, extra epochs are needed to ensure the weights are optimized effectively for the initial learning process.

\item \textbf{Outcome:} This helps create robust feature representations that can serve as a basis for learning subsequent tasks.
\end{itemize}
\textbf{2. CIFAR100 Experiments}

\begin{itemize}
    \item \textbf{Epochs for Task 1 : 25}
    
In CIFAR100 experiments, the first task receives 25 epochs. The larger number of epochs compared to MNIST reflects the increased complexity of the CIFAR100 dataset, which consists of higher-resolution images and more diverse classes.

\item \textbf{Reasoning:} The model requires more training time to process and extract meaningful patterns from the more complex data of the first 10 classes.

\item \textbf{Outcome:} This ensures that the model builds a robust understanding of the dataset, reducing the risk of misrepresentation as new tasks are introduced.

\end{itemize}

\subsubsection{Training Epochs for Remaining Tasks}
\textbf{1. MNIST Experiments}
\begin{itemize}
    \item \textbf{Epochs for Remaining Tasks: 5}

After the initial task, the model undergoes 5 epochs of training for each new task.

\item \textbf{Reasoning:} By this stage, the model has already developed basic feature extraction capabilities. The focus shifts to integrating new class-specific knowledge while retaining previously learned information.

\item \textbf{Efficiency:} Limiting the epochs ensures the process is computationally efficient, reducing training time without significantly compromising performance.
\end{itemize}

\textbf{2. CIFAR100 Experiments}
\begin{itemize}
    \item \textbf{Epochs for Remaining Tasks: 10}
    
Each subsequent task in the CIFAR100 experiments is trained for 10 epochs.

\item \textbf{Reasoning:} The complexity of CIFAR100 images necessitates more epochs per task compared to MNIST to fine-tune the model for new classes. However, it is fewer than Task 1 epochs since the initial learning has established general features that the model can reuse.

\item \textbf{Outcome:} This balanced approach prevents overfitting to new tasks while maintaining computational feasibility.
\end{itemize}

\subsubsection{Balancing Training Objectives}
The differences in epochs between Task 1 and the remaining tasks reflect the following priorities:

\begin{itemize}
    \item \textbf{Thorough Initial Learning:} Extra epochs for Task 1 ensure that the model builds a strong initial representation.

\item \textbf{Efficiency in Subsequent Tasks:} Reduced epochs for remaining tasks help balance computational resources and prevent overfitting to newer tasks while preserving knowledge from previous tasks.
\end{itemize}

This epoch distribution aligns with the CL goal of incrementally acquiring knowledge across tasks while mitigating forgetting, ensuring robust performance over sequential learning scenarios.

\textbf{Loss Function and Replay Loss Scaling:} The primary loss function used during training is the Cross-Entropy Loss, which is standard for classification tasks. This loss measures the divergence between the predicted probability distribution and the true class labels, ensuring the model learns accurate classifications. Additionally, for tasks beyond the first, a replay mechanism is employed to mitigate catastrophic forgetting. Replay loss, which incorporates data from previous tasks, is scaled by a factor of 0.5 to balance its contribution with the primary task's loss. This scaling ensures that while the model retains knowledge from earlier tasks, it does not overly prioritize old data at the expense of learning new information.

\textbf{Accuracy Metrics and Task-Level Evaluation:} The script calculates accuracy as the ratio of correctly predicted samples to the total number of samples in a batch. This metric is computed for each task and aggregated across all tasks learned so far. Incremental accuracy is determined as the mean of these task-specific accuracies, offering a comprehensive view of the model’s performance throughout its CL journey. Forgetting is assessed by comparing the accuracy on previous tasks before and after learning the current task. This metric quantifies the degree to which learning new tasks affects the model’s retention of earlier tasks.

\textbf{Batch Processing and Memory Optimization:} Training is performed in mini-batches, a standard practice that optimizes memory usage and accelerates computations on modern hardware like GPUs. This approach splits the dataset into manageable portions, allowing the model to iteratively update its parameters. Mini-batch training ensures that gradients are calculated and applied more frequently, leading to smoother and more stable convergence.


\textbf{Incremental Learning and Task-Specific Configurations:} Incremental learning strategies are embedded into the training process, where the model learns tasks sequentially while retaining knowledge from previous tasks. This is achieved through mechanisms like replay loaders and task-specific data loaders. Each task has its unique configuration for the number of epochs, replay data usage, and evaluation methods, ensuring the training process is tailored to the specific requirements of each task.

\subsection{Implementation details}
All the experiments have been performed on MNIST and CIFAR100 datasets by using NVIDIA RTX-3090 GPU and 3.8 GHz CPU with 64 GB RAM. The models are implemented using the PyTorch framework.

\section{Results and Discussion}
\label{Results}
The pursuit of adaptable and efficient machine learning models has led to the exploration of novel architectures capable of learning continuously without forgetting previously acquired knowledge. In this work, we delve into the efficacy of KANs in the context of CL for image classification, a paradigm shift from traditional machine learning that necessitates models to sequentially learn from a stream of tasks \cite{lesort2021understanding}. The challenge of catastrophic forgetting, wherein models experience a significant decline in performance on previously learned tasks upon learning new ones, is a central concern in CL \cite{gupta2025personalized}.  Our investigation encompasses a comparative analysis of standalone MLPs and KAN variants, alongside an examination of KAN's integration into the ViTs architecture \cite{parisigerman2019continual}. Through empirical evaluations on the MNIST and CIFAR100 datasets, we aim to elucidate the strengths and limitations of KAN in mitigating catastrophic forgetting and enhancing overall CL performance \cite{theotokis2025human}.

Our experimental design involved partitioning the MNIST and CIFAR100 datasets into distinct tasks to simulate a CL scenario, mirroring the real-world challenges of non-stationary data distributions. Specifically, the MNIST dataset, comprising 10 classes, was divided into 5 tasks, each containing 2 classes, while the CIFAR100 dataset, with its 100 classes, was split into 10 tasks, each encompassing 10 classes. The rationale behind this setup was to mimic a scenario where a model encounters new information incrementally, thereby requiring it to adapt and learn without compromising its existing knowledge base. The models underwent training for a predetermined number of epochs per task, with variations in the number of epochs for the initial task compared to subsequent tasks. This allowed us to observe how the models adapted to the initial learning phase versus subsequent incremental learning phases. We evaluated the performance of KAN variants against traditional MLPs and within the ViT architecture, focusing on metrics indicative of both forward learning (acquiring new knowledge) and backward transfer (retaining old knowledge). 

The core objective of this research was to rigorously assess the potential of KANs as a viable alternative to traditional MLPs in CL settings, with a specific focus on image classification tasks. By comparing the performance of KAN variants with traditional MLPs and integrating KAN into the ViT architecture, we aimed to gain insights into the potential advantages and drawbacks of using KAN for CL tasks, offering a comprehensive understanding of KAN's capabilities in this domain. Our findings suggest that, in standalone configurations, KAN models exhibit a notable advantage in CL scenarios, showcasing superior resistance to catastrophic forgetting compared to their MLP counterparts. The theoretical underpinnings of KANs lie in the Kolmogorov-Arnold Representation theorem, which posits that any multivariate continuous function can be expressed as a finite sum of compositions of continuous univariate functions. This inherent flexibility, coupled with learnable activation functions on edges, empowers KANs to approximate complex functions with greater precision compared to MLPs with fixed activation functions.

The integration of KANs into the ViT architecture yielded a nuanced set of results, with a slight improvement in average incremental accuracy observed, particularly during the initial stages of incremental learning, suggesting that KANs can facilitate faster adaptation to new tasks early in the learning process. However, as the CL process progressed, the performance of KAN-ViT converged with that of MLP-ViT, indicating that the benefits of KANs may diminish over time, possibly due to the increasing complexity of the learned representations or optimization challenges associated with deeper architectures. One potential explanation for this convergence is the inherent capacity of ViTs to capture long-range dependencies through self-attention mechanisms, which may overshadow the representational advantages offered by KANs in deeper layers. Further investigation is warranted to explore strategies for maximizing the benefits of KANs within transformer-based architectures for CL. The use of attention modules and large-scale pre-training contributes to ViT's robustness \cite{rahman2023out}.


The integration of KANs into ViT aims to enhance the model's capacity to adapt to new tasks without compromising its ability to perform CL. The architecture of KAN is based on B-splines, creating smooth curves through the usage of a set of control points. The result illustrates that the KAN-based ViTs achieve higher accuracy than traditional MLP-based ViTs by using learnable activation functions on edges. 

\subsection{Overall Performance Analysis}
The overall performance of the proposed technique has been performed by utilizing all the evaluation metrics. Table \ref{simpleresults} compares the performance of three models—FastKAN, MLP, and EfficientKAN—on the CIFAR100 and MNIST datasets, highlighting their accuracy, convergence speed, and computational efficiency. On CIFAR100, EfficientKAN achieves the highest test accuracy (57.5\%). FastKAN, while slightly less accurate on CIFAR100 (54.6\%), provides a faster training and testing time (1.939 s/epoch and 0.123 s/epoch, respectively). In contrast, MLP demonstrates poor performance on CIFAR100, with only 54.0\% test accuracy and the lowest train accuracy (64.0\%). On MNIST, FastKAN delivers the best overall performance with a test accuracy of 98.3\%, achieving a good balance between accuracy and computational efficiency. EfficientKAN closely follows with 98.1\% test accuracy, requiring fewer epochs (7) to converge but with the highest training (4.389 s/epoch) and testing (0.405 s/epoch) times. MLP, while achieving comparable test accuracy (98.0\%) but sacrifices generalization on more complex datasets like CIFAR100.

\begin{table*}[!ht]
\centering
\caption{Performance Comparison of FastKAN, MLP, and EfficientKAN.}
\scalebox{0.7}{
\begin{tabular}{ccccccc}
\hline
\textbf{Dataset}                                    & \textbf{Model} & \textbf{Epoch} & \textbf{\begin{tabular}[c]{@{}c@{}}Train Accuracy\\ (\%)\end{tabular}} & \textbf{\begin{tabular}[c]{@{}c@{}}Test Accuracy\\ (\%)\end{tabular}} & \textbf{\begin{tabular}[c]{@{}c@{}}Train Inference\\ Time Per\\ Epoch(s)\end{tabular}} & \textbf{\begin{tabular}[c]{@{}c@{}}Test Inference\\ Time Per\\ Epoch(s)\end{tabular}} \\ \hline
\multirow{3}{*}{\textbf{CIFAR100}}                  & FASTKAN        & 7              & 98.5                                                                   & 54.6                                                                  & 1.939                                                                                  & 0.123                                                                                 \\
                                                    & MLP            & 8              & 64.0                                                                   & 54.0                                                                  & 1.125                                                                                  & 0.043                                                                                 \\
                                                    & EfficientKAN   & 9              & 86.3                                                                   & \textbf{57.5}                                                         & 3.663                                                                                  & 0.326                                                                                 \\ \hline
\multicolumn{1}{l}{\multirow{3}{*}{\textbf{MNIST}}} & FastKAN        & 10             & 99.9                                                                   & \textbf{98.3}                                                         & 2.331                                                                                  & 0.117                                                                                 \\
\multicolumn{1}{l}{}                                & MLP            & 9              & 99.0                                                                   & 98.0                                                                  & 1.341                                                                                  & 0.025                                                                                 \\
\multicolumn{1}{l}{}                                & EfficientKAN   & 7              & 99.9                                                                   & 98.1                                                                  & 4.389                                                                                  & 0.405                                                                                 \\ \hline
\end{tabular}
}
\label{simpleresults}
\end{table*}

Overall, EfficientKAN excels in accuracy, particularly for more complex datasets, while FastKAN provides a balanced trade-off between accuracy and efficiency. The results indicate that the model choice depends on the specific dataset and whether accuracy or computational speed is prioritized.

\begin{table}[!ht]
\centering
\caption{CL Performance Comparison: Simple MLP vs. EfficientKAN}
\scalebox{0.8}{
\begin{tabular}{ccccc}
\hline
\textbf{Model}        & \textbf{Dataset} & \textbf{\begin{tabular}[c]{@{}c@{}}Average Incremental\\ Accuracy(\%)\end{tabular}} & \textbf{\begin{tabular}[c]{@{}c@{}}Last Task\\ Accuracy(\%)\end{tabular}} & \textbf{\begin{tabular}[c]{@{}c@{}}Average Global\\ Forgetting\end{tabular}} \\ \hline
\textbf{MLP}          & MNIST            & 45.8                                                                                & 22.1                                                                      & 95.2                                                                        \\
\textbf{EfficientKAN} & MNIST            & \textbf{52.2}                                                                       & \textbf{50.9}                                                             & \textbf{71.0}                                                               \\ \hline
\end{tabular}
}
\label{CL}
\end{table}

Table \ref{CL} presents a comparison of CL performance metrics between the MLP and EfficientKAN models on the MNIST dataset, focusing on three key indicators: Average Incremental Accuracy, Last Task Accuracy, and Average Global Forgetting. EfficientKAN outperforms MLP across all metrics. It achieves a higher Average Incremental Accuracy of 52.2\% compared to MLP's 45.8\%, indicating that EfficientKAN consistently performs better across sequential tasks. For Last Task Accuracy, EfficientKAN significantly surpasses MLP, achieving 50.9\% compared to MLP's 22.1\%. This result highlights EfficientKAN's superior ability to adapt to new tasks while retaining knowledge from earlier tasks. In terms of Average Global Forgetting, EfficientKAN demonstrates a marked improvement with a value of 71.0\%, significantly lower than MLP's 95.2\%. Lower forgetting indicates that EfficientKAN is more effective at preserving knowledge from earlier tasks in a CL setting.

In general, these results emphasize EfficientKAN's robustness in mitigating catastrophic forgetting and its superior adaptability and retention in CL scenarios compared to MLP.

Table \ref{vitbase} compares the CL performance of ViT-MLP and ViT-KAN across two datasets: MNIST and CIFAR100, focusing on three key metrics. For the MNIST dataset, ViT-KAN achieves a slightly higher Average Incremental Accuracy (18.44\%) compared to ViT-MLP (17.70\%), indicating better performance in handling sequential tasks. However, in Last Task Accuracy, ViT-MLP (6.66\%) outperforms ViT-KAN (4.67\%), suggesting that ViT-MLP retains better accuracy for the most recent task. In terms of Average Global Forgetting, ViT-KAN demonstrates a marginal improvement with 33.47\% compared to ViT-MLP's 35.16\%, reflecting better retention of earlier tasks. For the CIFAR100 dataset, ViT-KAN again shows superior Average Incremental Accuracy (15.49\%) over ViT-MLP (13.63\%), underscoring its effectiveness in sequential task learning. In Last Task Accuracy, both models perform almost identically, with ViT-KAN scoring 4.77\% and ViT-MLP scoring 4.75\%. However, in Average Global Forgetting, ViT-KAN exhibits slightly higher forgetting (49.85\%) compared to ViT-MLP (44.67\%), indicating a trade-off in retention for earlier tasks in this case.

\begin{table}[!ht]
\centering
\caption{Performance comparison of CL of ViT\_KAN vs ViT\_MLP tasks.}
\begin{tabular}{ccccc}
\hline
\textbf{Dataset}          & \textbf{Model} & \textbf{\begin{tabular}[c]{@{}c@{}}Average \\ Incremental\\ Accuracy\end{tabular}} & \textbf{\begin{tabular}[c]{@{}c@{}}Last Task\\ Accuracy\end{tabular}} & \textbf{\begin{tabular}[c]{@{}c@{}}Average\\ Global Forgetting\end{tabular}} \\ \hline
\multirow{2}{*}{MNIST}    & ViT\_MLP       & 17.70                                                                             & 6.66                                                                & 35.16                                                                       \\
                          & ViT\_KAN       & \textbf{18.44}                                                                    & 4.67                                                                & 33.47                                                                       \\ \hline
\multirow{2}{*}{CIFAR100} & ViT\_MLP       & 13.63                                                                             & 4.75                                                                & 44.67                                                                       \\
                          & ViT\_KAN       & \textbf{15.49}                                                                    & 4.77                                                                & 49.85                                                                       \\ \hline
\end{tabular}
\label{vitbase}
\end{table}

The results highlight ViT-KAN as a promising alternative to ViT-MLP, with noticeable improvements in incremental learning accuracy and modest benefits in catastrophic forgetting, especially on MNIST. However, the trade-offs in retention and last-task accuracy on CIFAR100 suggest areas for further refinement.

Table \ref{replay} presents a performance comparison between ViT-MLP and ViT-KAN in the context of CL on the CIFAR100 dataset, incorporating replay mechanisms. ViT-KAN outperforms ViT-MLP in Average Incremental Accuracy, achieving 17.23\% compared to ViT-MLP's 16.58\%. This indicates that ViT-KAN is better at learning sequential tasks while maintaining a higher overall performance across all tasks. In Last Task Accuracy, which reflects the model's ability to retain performance on the most recent task, ViT-KAN again has an edge, scoring 6.47\% compared to ViT-MLP's 6.09\%. This suggests that ViT-KAN adapts more effectively to new tasks while preserving recent knowledge. However, in terms of Average Global Forgetting Accuracy, which measures the overall forgetting across all tasks, ViT-KAN shows slightly higher forgetting (57.45\%) compared to ViT-MLP (55.72\%). While this represents a trade-off, the increased incremental and last-task accuracies suggest that ViT-KAN prioritizes learning and adapting to new tasks at a small cost to knowledge retention from earlier tasks.

\begin{table}[!ht]
\centering
\caption{CL Performance: ViT\_KAN vs. ViT\_MLP with Replay.}
\begin{tabular}{ccccc}
\hline
\textbf{Model}    & \textbf{Dataset} & \textbf{\begin{tabular}[c]{@{}c@{}}Average\\ Incremental\\ Accuracy\end{tabular}} & \textbf{\begin{tabular}[c]{@{}c@{}}Last Task\\ Accuracy\end{tabular}} & \textbf{\begin{tabular}[c]{@{}c@{}}Average\\ Global Forgetting\\ Accuracy\end{tabular}} \\ \hline
\textbf{ViT\_MLP} & CIFAR100         & 16.58                                                                            & 6.09                                                                & 55.72                                                                                  \\
\textbf{ViT\_KAN} & CIFAR100         & \textbf{17.23}                                                                   & \textbf{6.47}                                                       & 57.45                                                                                  \\ \hline
\end{tabular}
\label{replay}
\end{table}

In conclusion, the inclusion of replay mechanisms highlights ViT-KAN as a more effective model for CL on CIFAR100, balancing better task adaptation and incremental performance with slightly increased global forgetting.

\begin{figure*}[!ht]
     \centering
     \includegraphics[width=1\textwidth]{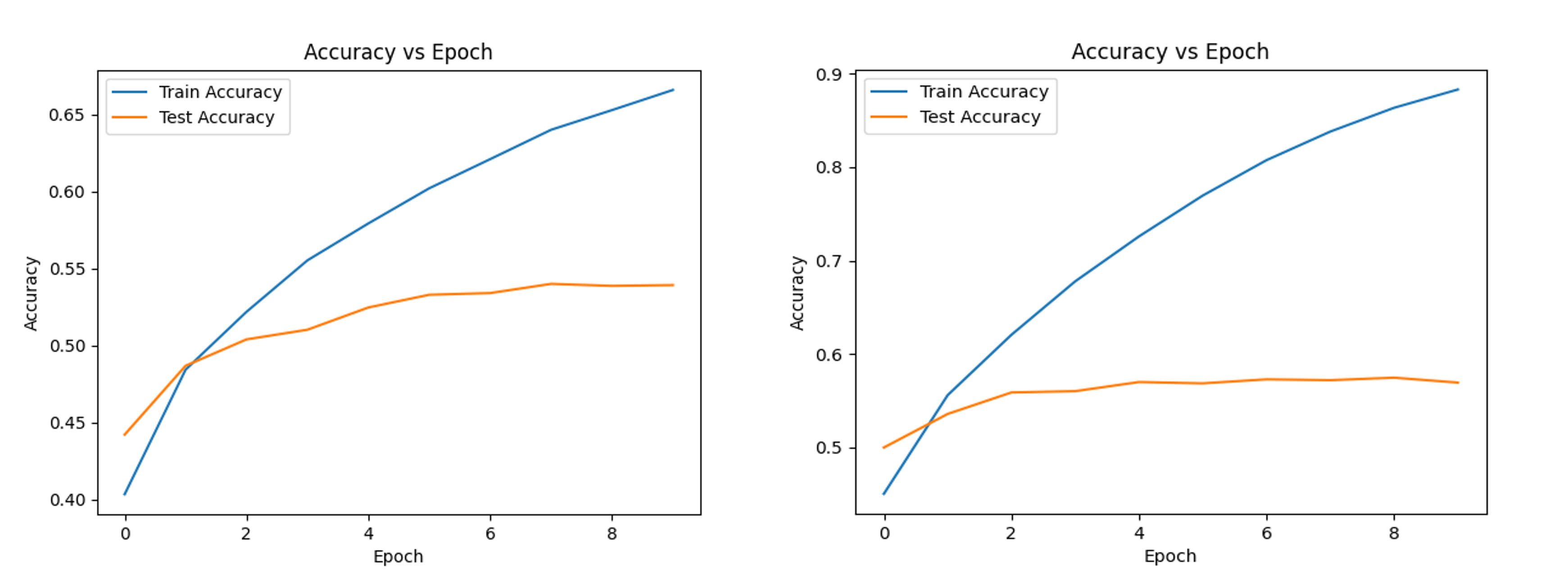}
     \caption{Base level accuracy of MLP and efficientKAN on CIFAR-100 dataset. }
     \label{MLP_Eff-KAN_CIFAR}
 \end{figure*}

Fig. \ref{MLP_Eff-KAN_CIFAR} provides a comparative analysis of the training and testing performance of MLP and EfficientKAN models on the CIFAR-100 dataset in terms of accuracy. For both models, train accuracy steadily increases as the number of epochs progresses, indicating that both models effectively learn from the training data. However, the test accuracy of EfficientKAN is consistently higher than that of the MLP model. This suggests that EfficientKAN generalizes better to unseen data compared to the MLP model, as evidenced by the more significant gap in test accuracy improvement.

\begin{figure*}[!ht]
     \centering
     \includegraphics[width=1\textwidth]{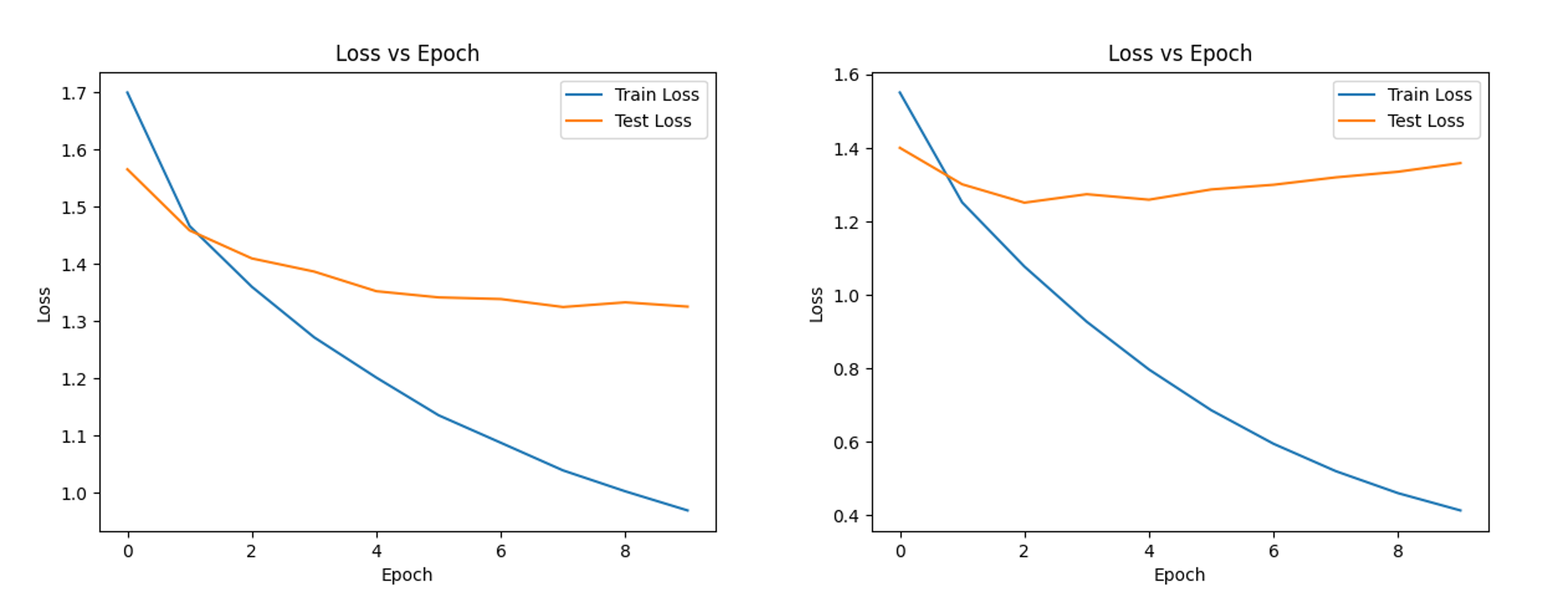}
     \caption{Base level loss of MLP and efficientKAN on CIFAR-100 dataset. }
     \label{MLP_Eff-KAN-loss_CIFAR}
 \end{figure*}

Fig. \ref{MLP_Eff-KAN-loss_CIFAR} shows the loss of MLP and EfficientKAN over multiple epochs. The loss curves demonstrate the convergence behavior for both models. The MLP shows a gradual decrease in train loss, but the test loss remains relatively higher, indicating potential overfitting. In contrast, the EfficientKAN reveals a sharper reduction in train loss alongside a significantly lower and more stable test loss, suggesting better learning dynamics and improved generalization to test data.

It can be observed from Fig. \ref{MLP_Eff-KAN_CIFAR} and \ref{MLP_Eff-KAN-loss_CIFAR} that EfficientKAN outperforms MLP in terms of both test accuracy and test loss. The results highlight EfficientKAN's ability to achieve superior generalization, likely due to its architectural advantages that promote efficient learning and robustness in CL scenarios. These observations underscore EfficientKAN's suitability for challenging tasks like CIFAR-100, where maintaining generalization is critical.

\begin{figure*}[!ht]
     \centering
     \includegraphics[width=1\textwidth]{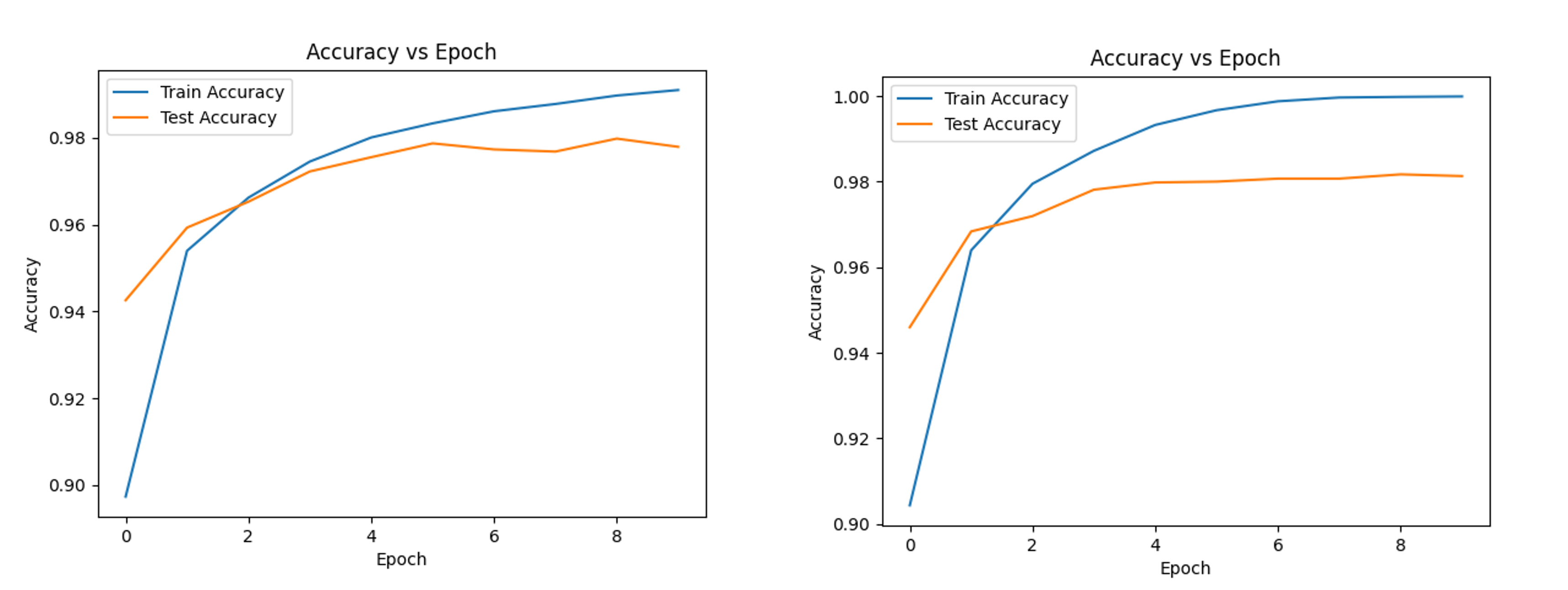}
     \caption{Base level accuracy of MLP and efficientKAN on MNIST dataset. }
     \label{MLP_Eff-KAN-MNIST}
 \end{figure*}

\begin{figure*}[!ht]
     \centering
     \includegraphics[width=1\textwidth]{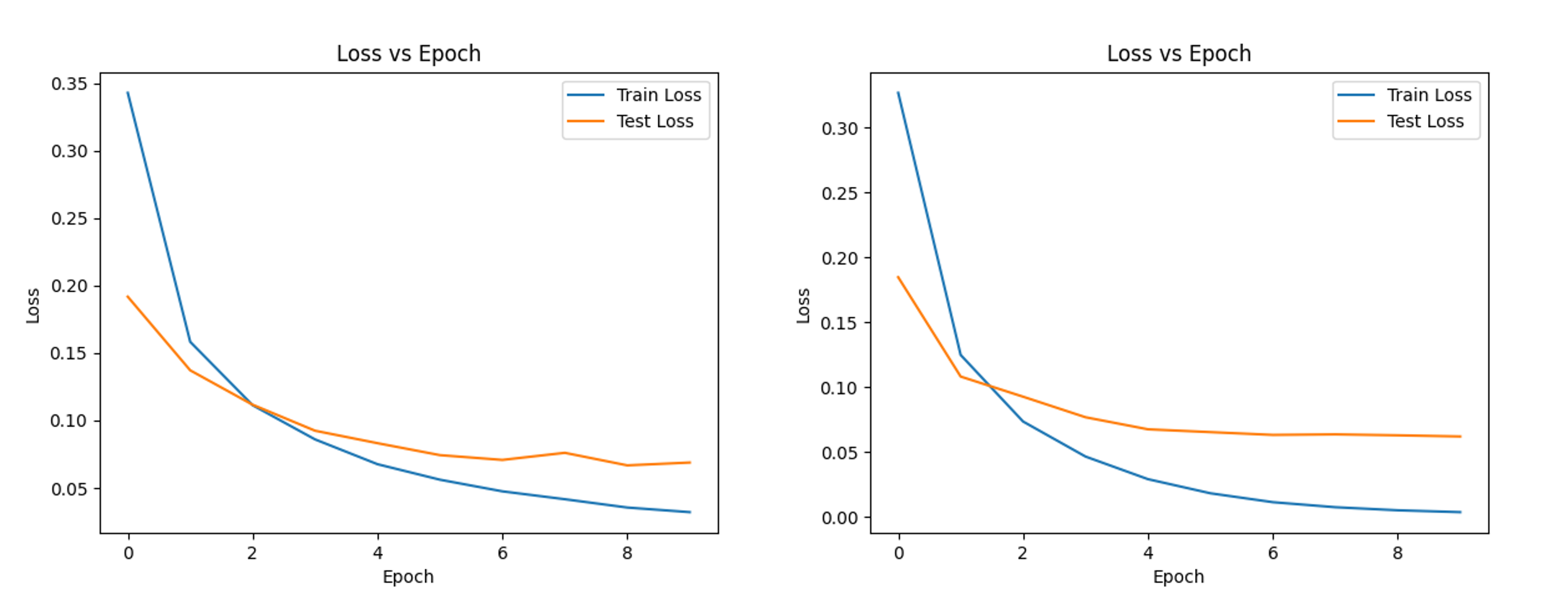}
     \caption{Base level loss of MLP and efficientKAN on MNIST dataset. }
     \label{MLP_Eff-KAN-loss_MNIST}
 \end{figure*}

Figures \ref{MLP_Eff-KAN-MNIST} and \ref{MLP_Eff-KAN-loss_MNIST} present the comparative performance of MLP and EfficientKAN models on the MNIST dataset, evaluating their accuracy and loss over training epochs.

In Figure \ref{MLP_Eff-KAN-MNIST}, both models show rapid improvement in train and test accuracy during the initial epochs, with the training accuracy nearing 100\%. However, EfficientKAN consistently achieves slightly higher test accuracy compared to MLP throughout the epochs. This suggests that EfficientKAN generalizes better to unseen data, maintaining its performance across training iterations.

Fig. \ref{MLP_Eff-KAN-loss_MNIST} illustrates the training and test loss curves for the two models. Both models exhibit a sharp reduction in training loss within the first few epochs, indicating effective convergence. The test loss, however, is consistently lower for EfficientKAN compared to MLP, demonstrating better generalization and stability during training. Additionally, the test loss curve for EfficientKAN shows less fluctuation, highlighting its reliability.

To conclude, EfficientKAN outperforms MLP in terms of both accuracy and loss on the MNIST dataset. It converges faster, achieves higher test accuracy, and exhibits lower test loss, showcasing its superior ability to extract and retain essential features while minimizing overfitting.

\section{Conclusion and Future Work}
\label{conclusion}
This study presents a novel approach to addressing the critical challenge of catastrophic forgetting in ViTs by replacing traditional MLP layers with KANs. Leveraging the local plasticity of spline-based activations in KANs, the proposed architecture demonstrates superior knowledge retention and adaptability in CL settings. Rigorous experiments on benchmark datasets such as MNIST and CIFAR100 reveal that KAN-based ViTs consistently outperform their MLP-based counterparts, showcasing improved task accuracy and resistance to catastrophic forgetting. A comprehensive analysis of optimization techniques highlights the trade-offs between first-order optimizers, with ADAM providing computational efficiency. This analysis emphasizes the critical role of optimizer selection in maximizing the performance of KAN-based ViTs. By introducing a systematic framework for sequential learning, this work provides valuable insights into the potential of KANs to enhance the scalability and robustness of ViTs in dynamic environments. These findings not only pave the way for more effective neural network architectures in CL scenarios but also open up avenues for further exploration of spline-based methods in other domains. In conclusion, the integration of KANs into ViTs represents a promising step toward creating more adaptable and efficient vision models, with significant implications for real-world applications requiring CL and robust knowledge retention.

Future research will explore the scalability and robustness of KAN-based ViTs across diverse datasets and CL benchmarks. Architectural innovations, such as dynamic spline activation functions and hierarchical KAN layers, will be examined to enhance model performance. Advanced regularization techniques, including elastic weight consolidation and sparsity constraints, will be integrated to further address catastrophic forgetting. Expanding the framework to multi-modal tasks and dynamic data streams will provide valuable insights into its real-world adaptability. Additionally, optimizing resource-efficient training methods will ensure practical feasibility and broader applicability.

\section*{CRediT authorship contribution statement}
\textbf{Zahid Ullah:} Conceptualization, Data curation, Methodology, Software, Formal analysis, Investigation, Writing - original draft, Writing - review \& editing. \textbf{Jihie Kim:} Conceptualization, Writing – review \& editing, Formal analysis, Investigation, Supervision, Project administration, Funding acquisition.

\section*{\textbf{Declaration of Competing Interests}} The authors declare that they have no known competing financial interests or personal relationships that could have appeared to influence the work reported in this paper.

\section*{Acknowledgements}
This research was supported by the MSIT(Ministry of Science and ICT), Korea, under the ITRC(Information Technology Research Center) support program(IITP-2025-RS-2020-II201789), and the Artificial Intelligence Convergence Innovation Human Resources Development(IITP-2025-RS-2023-00254592) supervised by the IITP(Institute for Information \& Communications Technology Planning \& Evaluation).

\bibliographystyle{elsarticle-num-names}
\bibliography{sample.bib}







\end{document}